\documentstyle[aclap]{article}

\makeatletter

\def\evnup{\@ifnextchar[{\@evnup}{\@evnup[0pt]}}

\def\@evnup[#1]#2{\setbox1=\hbox{#2}%
\dimen1=\ht1 \advance\dimen1 by -.5\baselineskip%
\advance\dimen1 by -#1%
\leavevmode\lower\dimen1\box1}

\def\fd#1{\setlength{\baselineskip}{0pt}
  \small
     \hbox{#1}}
\def\fdx#1{\setlength{\baselineskip}{0pt}\vcenter{#1}}
\def\fdand#1{\(\left[\fdx{#1}\right]\)}

\def\feat#1#2{\vskip .4ex\hbox{\hspace{.2em}#1\hspace{1em}#2\hspace{.2em}}\vskip .8ex}
\def\etc{\hbox to 1in{\hss\vdots\hss}}

\def\line#1#2(#3){\@stepcounter{equation}
  \let\@currentlabel=\theequation \label{#3}
  \hbox to \hsize{\hbox to #1in{(\theequation) \hfil}
       \hskip 0pt minus 0.4in $#2$\hfil}}

\makeatother

\title{Probabilistic Coreference in Information Extraction}
\author{Andrew Kehler \\ SRI International \\
 333 Ravenswood Avenue \\ Menlo Park, CA 94025 \\ kehler@ai.sri.com} 

\begin{document}
\bibliographystyle{fullname}
\maketitle

\begin{abstract}
Certain applications require that the output of an information
extraction system be probabilistic, so that a downstream system can
reliably {\it fuse} the output with possibly contradictory information
from other sources.  In this paper we consider the problem of
assigning a probability distribution to alternative sets of
coreference relationships among entity descriptions.  We present the
results of initial experiments with several approaches to estimating
such distributions in an application using SRI's FASTUS information
extraction system.
\end{abstract}

\section{Introduction}

Natural language information extraction (IE) systems take texts
containing natural language as input and produce database templates
populated with information that is relevant to a particular
application.  These records may be fed as input to a downstream system
for which the IE system is only one of several sources of information.
In such a scenario, the downstream system must {\it fuse} the incoming
information from each of its sources, requiring the resolution of
conflicts.  To accomplish this, the fusion system must know the
reliability of the information received from each source; in this way
unreliable information from one source can be disregarded in favor of
highly reliable information from another.

\setlength{\unitlength}{0.00083300in}%
\begingroup\makeatletter\ifx\SetFigFont\undefined
\def\x#1#2#3#4#5#6#7\relax{\def\x{#1#2#3#4#5#6}}%
\expandafter\x\fmtname xxxxxx\relax \def\y{splain}%
\ifx\x\y   
\gdef\SetFigFont#1#2#3{%
  \ifnum #1<17\tiny\else \ifnum #1<20\small\else
  \ifnum #1<24\normalsize\else \ifnum #1<29\large\else
  \ifnum #1<34\Large\else \ifnum #1<41\LARGE\else
     \huge\fi\fi\fi\fi\fi\fi
  \csname #3\endcsname}%
\else
\gdef\SetFigFont#1#2#3{\begingroup
  \count@#1\relax \ifnum 25<\count@\count@25\fi
  \def\x{\endgroup\@setsize\SetFigFont{#2pt}}%
  \expandafter\x
    \csname \romannumeral\the\count@ pt\expandafter\endcsname
    \csname @\romannumeral\the\count@ pt\endcsname
  \csname #3\endcsname}%
\fi
\fi\endgroup
\begin{figure*}[ht]
\begin{center}
\begin{picture}(5771,2499)(2200,-2398)
\thicklines
\put(2031,-1126){\vector( 1, 0){1000}}
\put(2200,-1000){\makebox(0,0)[lb]{\smash{\SetFigFont{8}{9.6}{rm}NL Text}}}
\put(4355,-2386){\framebox(1326,435){}}
\put(4466,-2125){\makebox(0,0)[lb]{\smash{\SetFigFont{9}{10.8}{rm}INFORMATION}}}
\put(4692,-2299){\makebox(0,0)[lb]{\smash{\SetFigFont{9}{10.8}{rm}SOURCE}}}
\put(3195,-1344){\framebox(1050,303){}}
\put(3207,-1300){\makebox(0,0)[lb]{\smash{\SetFigFont{8}{9.6}{rm}RECOGNITION}}}
\put(3361,-1171){\makebox(0,0)[lb]{\smash{\SetFigFont{8}{9.6}{rm}PATTERN}}}
\put(4632,-1344){\framebox(939,303){}}
\put(4772,-1300){\makebox(0,0)[lb]{\smash{\SetFigFont{8}{9.6}{rm}MERGING}}}
\put(4727,-1171){\makebox(0,0)[lb]{\smash{\SetFigFont{8}{9.6}{rm}TEMPLATE}}}
\put(4245,-1214){\vector( 1, 0){387}}
\put(3029,-1518){\framebox(2708,825){}}
\put(3305,-909){\makebox(0,0)[lb]{\smash{\SetFigFont{9}{10.8}{rm}INFORMATION EXTRACTION}}}
\put(6787,-1518){\framebox(1989,825){}}
\put(7229,-1083){\makebox(0,0)[lb]{\smash{\SetFigFont{9}{10.8}{rm}DOWNSTREAM }}}
\put(7284,-1257){\makebox(0,0)[lb]{\smash{\SetFigFont{9}{10.8}{rm}PROCESSING}}}
\put(5626,-128){\vector( 3,-2){1144.385}}
\put(5737,-1126){\vector( 1, 0){1050}}
\put(5900,-1000){\makebox(0,0)[lb]{\smash{\SetFigFont{8}{9.6}{rm}Templates}}}
\put(4300,-346){\framebox(1326,435){}}
\put(4410,-84){\makebox(0,0)[lb]{\smash{\SetFigFont{9}{10.8}{rm}INFORMATION}}}
\put(4636,-258){\makebox(0,0)[lb]{\smash{\SetFigFont{9}{10.8}{rm}SOURCE}}}
\put(5681,-2213){\vector( 3, 2){1106.769}}
\end{picture}
\end{center}
\label{arch-figure}
\caption{A Scenario Employing an Information Extraction System}
\end{figure*}
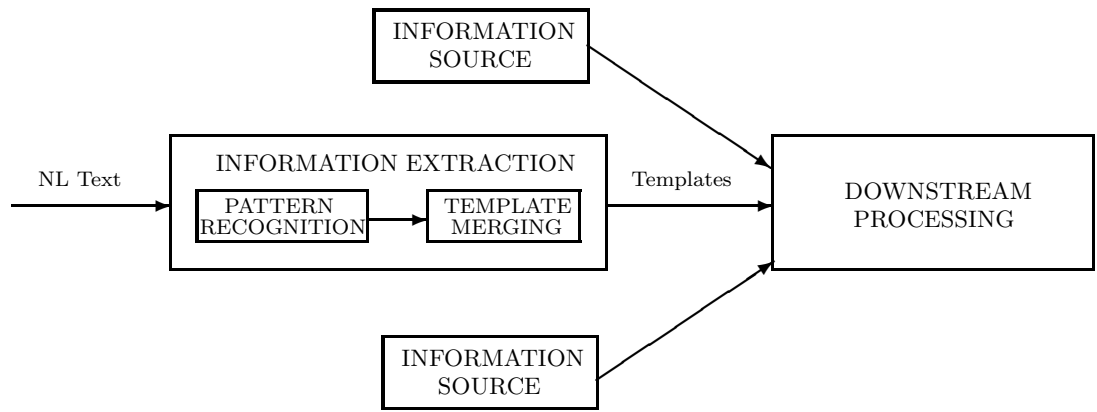

Figure~1 exhibits this scenario with a typical IE system such as SRI's
FASTUS system \cite{HobbsEtAl:96a}.  The IE system has two components.
The first component consists of a series of phases that recognize
domain-relevant patterns in the text and create templates representing
event and entity descriptions from them.  The second component merges
templates created from different phrases in the text that overlap in
reference.  The resulting set of templates constitutes a formal
description of the state of affairs as described in the text with
respect to the application specification, which is then fed to the
downstream system.

As part of determining this state of affairs, the IE system must
create templates describing the relevant entities that are reported
on.  This requires determining when two or more templates describe the
same entity, as templates created from coreferring phrases need to be
merged.  We have performed an informal study of FASTUS's processing of
a set of texts which indicates that the merging phase is where most of
the ambiguities (as well as most of the errors) lie.  However, most IE
systems, including FASTUS, have pursued a deterministic strategy for
merging and report only a single possible state of affairs.  This
limitation makes it difficult for a downstream system to fuse the
information with possibly contradictory information from other
sources, as no information about the IE system's certainty of the
results is passed along, nor is information about possible alternative
states of affairs and their associated levels of certainty.

In this paper, we consider the problem of assigning a probability
distribution to alternative sets of coreference relationships among
entity descriptions.  We present the results of initial experiments
with several approaches to estimating such distributions in an
application using FASTUS.

\section{Overview of the Problem}
\label{problem-section}

Let us consider an example text of the sort that we encounter 
in our application:\footnote{The texts in our application are messages
consisting of free text, possibly interspersed with formatted tables
or charts which themselves may contain natural language fragments that
require analysis.  While this example is shorter than most texts in
our corpus, the relevant free text portions of the messages are
typically no longer than a few paragraphs.  The style displayed in this
example is fairly typical, although in some cases the sentence
structure is more telegraphic.}

\begin{quote}
Subj: {\it Kinston Military Rail Depot} \vspace*{.1in}

{\it A rail depot} was found 100 km southwest of the capitol of
Raleigh, consisting of extensive admin and support areas (similar to
{\it the ammunition depot in Fairview}), two material storage areas,
extensive transshipment facilities (some of which are under
construction immediately east of {\it the depot}), and several
training areas.
\end{quote}

We focus on the four mentions of depots in the text, which are
highlighted with italics.  The pattern matching phases of FASTUS
produce templates similar to those shown in Figure~2.

\begin{figure}[ht]
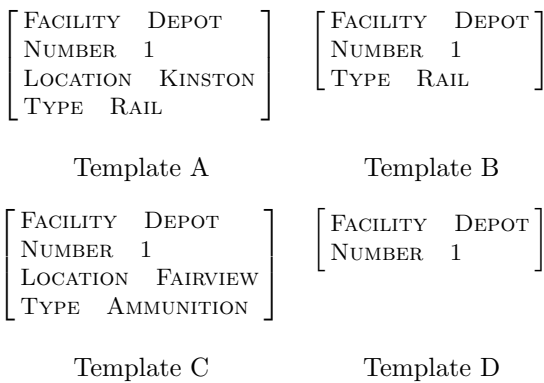

\begin{center}
\begin{tabular}{cc} 

\hspace*{-.2in}

\evnup{
\fd{
\fdand{\feat{\sc Facility}{\sc Depot}
	\feat{\sc Number}{\sc 1}
	\feat{\sc Location}{\sc Kinston}
	\feat{\sc Type}{\sc Rail}}}}
&
\hspace*{-.2in}

\evnup{
\fd{
\fdand{\feat{\sc Facility}{\sc Depot}
	\feat{\sc Number}{\sc 1}
	\feat{\sc Type}{\sc Rail}}}} \\ \\
Template A & Template B \\ 
\hspace*{-.2in}
\evnup{
\fd{
\fdand{\feat{\sc Facility}{\sc Depot}
	\feat{\sc Number}{\sc 1}
        \feat{\sc Location}{\sc Fairview}
	\feat{\sc Type}{\sc Ammunition}}}}
& 
\hspace*{-.2in}

\evnup{
\fd{
\fdand{\feat{\sc Facility}{\sc Depot}
	\feat{\sc Number}{\sc 1}}}} \\ \\ 
Template C & Template D
\end{tabular}
\end{center}
\caption{Templates Representing Depots Mentioned}
\label{templates}
\end{figure}
We will refer to a set of templates that have potential coreference
relationships among them as a {\it coreference
set},\footnote{Templates A, B, C, and D constitute the only
coreference set in this example, since none of the other NPs (e.g.,
the various ``areas'' mentioned) are compatible with any of the
others.  In general, however, a text can give rise to any number of
distinct coreference sets, each of which will be assigned its own
probability distribution.}  and possible partitions of coreferential
templates in the set as {\it coreference configurations}.  In the
coreference set containing templates A, B, C, and D, system knowledge
external to the probabilistic model indicates that the type {\it
Ammunition} in template C is not compatible with the type {\it Rail}
in A and B; therefore these are taken {\it a priori} to be
non-coreferential.  Given these incompatibilities, seven possible
coreference configurations remain.  Template names grouped within
parentheses are taken to be mutually coreferring; we will refer to
such a grouping as a {\it cell} of the coreference configuration.

\begin{center}
\begin{tabular}{l l} 
1. (A B D) (C)  & 5. (B D) (A) (C) \\
2. (A B) (C D)  &  6. (C D) (A) (B) \\
3. (A B) (C) (D) & 7. (A) (B) (C) (D) \\
4. (A D) (B) (C) & \\
\end{tabular}
\end{center}
The first of these configurations expresses the correct coreference
relationships for the example.

Given a coreference set of templates, possibly coupled with a list of
template pairs known a priori not to corefer, the task is to assign a
probability distribution over the possible coreference configurations
for that set.

\paragraph{Relationship to Past Work} 
While there have been previous investigations of empirical approaches
to coreference, these have generally centered on the task of assigning
correct referents for {\it anaphoric} expressions
\cite{ConnollyEtAl:94,AoneBen:95,LappinLeass:94,DaganItai:90,DaganEtAl:95,KennedyBoguraev:96a,KennedyBoguraev:96b}.
The current task deviates from that problem in several respects.
First, in our task, all coreference relationships among templates are
modeled regardless of the ``referentiality'' of the phrases that led
to their creation.  For instance, indefinites will sometimes corefer
with a previously described entity; a typical case is illustrated by
the coreference between the indefinite ``a rail depot'' and the
depot introduced in the subject line in the example passage.  Also,
entities described with bare plurals are commonly found to be
coreferential with other entities, in addition to cases in which they
have their more standard generic meanings.  On the other hand,
definite noun phrases are often not referential to items evoked in the
text (e.g., ``the ammunition depot in Fairview'').  Determining when
such expressions are discourse-anaphoric is part of the task; this
information is generally not known to the system a priori.

Second, the results of this task will be evaluated by the probability
assigned to the correct state of affairs with respect to an entire
coreference set, and not by the number of correct antecedents assigned
to anaphoric expressions.  Modeling at the level of coreference sets
ensures that the probabilities are consistent when
considering the global state of affairs being described in the
text.  Furthermore, the role of probabilities for this application
goes beyond selecting the correct coreference relationships -- the
probability assigned to an alternative will be central in determining
how the downstream system will weigh it against information from other
sources during data fusion.  A system that assigns a probability of
0.9 to correct answers is more successful than one that assigns a
probability of 0.6 to them.

\paragraph{The Limitations of IE Systems} 
The properties of typical IE systems such as FASTUS also make this
task challenging.  For one, successful modeling of coreference
relationships is hampered by the crudeness of the representations
used.  The templates that are created are fairly shallow and may be
incomplete.  A reliance on detailed information about the context can
prove detrimental if such information is often missed by the system.
Also, FASTUS also does not build up complex representations for the
syntax and semantics of sentences, placing limits on the extent to
which such information can be utilized in determining coreference.
Lastly, there are the inaccuracies that result from processing real
text.  The pattern matching phases of FASTUS may intermittently
misanalyze phrases that serve as antecedents for subsequent referring
expressions.  Therefore, for example, with respect to an identified
coreference set, it may be correct to place a referential pronoun in
its own cell (implying that it does not corefer with anything), simply
because system error caused its antecedent not to be included in the
set.

\paragraph{Outline of the Approach}
The number of coreference configurations over which 
a distribution is to be assigned depends on the number of templates in
the coreference set, and the set of a priori constraints against
coreference between some of its members.  As there are many
scenarios that will never be encountered in a corpus of training data of
any reasonable size, it would be hopeless to attempt to estimate a
conditional distribution for each possibility directly.
To make matters worse, training data comes at a cost, as
keys have to be coded by hand.  One of the goals of this effort
is to allow the ability to train up probabilities in new domains
quickly, which requires an approach that is successful with a limited
amount of training data.

However, it would be reasonable to expect that we have enough data to
estimate distributions for coreference sets with only two members.
This suggests a two-step approach.  First, we develop a general model
of coreference between any two templates, and apply it to 
pairwise combinations of templates in a given coreference set
without regard to the other templates in the set.  We then utilize
a method for combining the resulting probabilities to form a
distribution over all the possible coreference configurations.  We
describe our method for modeling probabilities between pairs of
templates in the next section, and describe two methods for deriving a
distribution over the coreference configurations in
Section~4.  We report on an evaluation and
comparison of the approaches in Section~5. 

\section{Training A Model for Pairs of Templates}

Our first task is to derive a model for determining the probability
that two templates corefer, conditioned on various characteristics of
the context.  For this we employ an approach to maximum entropy
modeling described by Berger et
al. \shortcite{BergerEtAl:96}.

\paragraph{Maximum Entropy Modeling}

Suppose we wish to model some random process, such as that which
determines coreference between two templates generated by an IE
system, based on various characteristics of the context that influence
this process, such as the content of the templates themselves, the
form of the natural language expressions from which the templates were
created, and the distance between those expressions in the text.  We
refer to the collection of such characteristics for a given example as
its context $x$, and the value denoting the output of the process as
$y$.  We can define a set of binary {\it features} that relate a
possible value of a characteristic of $x$ with a possible outcome $y$,
i.e., whether the two templates corefer ($y=1$) or not ($y=0$).  For
example, a feature $f_1(x,y)$ pairing the characteristic of S and T
having identical slot values with the outcome that they corefer would
be defined as follows.

\vspace*{.17in}

\noindent {\bf Binary Feature f$_1$(x,y)}: \\

\vspace*{.11in}

$f_1(x,y)$ = $\left\{ \begin{tabular}{l}
1 if S and T have identical  \\ \hspace*{.11in} slot values and S
and T corefer \\
0 otherwise
\end{tabular} \right.$

\vspace*{.17in}
From these features we can define {\it constraints} on the
probabilistic model that is learned, in which we assume that the
expected value of the feature with respect to the distribution of the
training data ($p_d$) holds with respect to the general model ($p_m$).

\vspace*{.15in}

\noindent {\bf Constraints}: 

\vspace*{.11in}

\[ \sum_{x,y} p_{d}(x,y) f(x,y) = \sum_{x,y} p_{d}(x) p_{m}(y|x)
f(x,y) \] 
Given that we have chosen a set of such constraints to impose on our
model, we wish to identify that model which has the maximum entropy --
this is the model that assumes the least information beyond those
constraints.  Berger et al. \shortcite{BergerEtAl:96} show that this
model is a member of an exponential family with one parameter for each
constraint, specifically a model of the form

\[ p(y|x)= \frac{1}{Z(x)} e^{\sum_{i} \lambda_{i} 
f_{i}(x,y)} \]
in which

\[ Z(x) = \sum_{y} e^{\sum_{i} \lambda_{i} f_{i}(x,y)} \] 
The parameters $\lambda_1,...,\lambda_n$ are Lagrange multipliers that
impose the constraints corresponding to the chosen features
$f_1,...,f_n$.  The term $Z(x)$ normalizes the probabilities by
summing over all possible outcomes $y$.  Berger et
al. \shortcite{BergerEtAl:96} demonstrate that the optimal values for
the $\lambda_{i}$'s can be obtained by maximizing the likelihood of
the training data with respect to the model, which can be performed
using their {\it improved iterative scaling} algorithm.

In practice, we will not want to incorporate constraints for all of
the features that we might define, but only those that are most
relevant and informative.  Therefore, we use a procedure for selecting
which of our pool of features should be made {\it active}.  At each
iteration, the algorithm approximates the gain in the model's
predictiveness that would result from imposing the constraints
corresponding to each of the existing inactive features, and selects
the one with the highest anticipated payoff.  Upon making this feature
active, the $\lambda_{i}$'s for all active features are
(re)trained so that the constraints are all met simultaneously.  The
feature selection process is iterated until the approximate gain for
all the remaining inactive features is negligible.

\paragraph{Characteristics of Context for Template Coreference} 
We now need a set of possible characteristics of context
on which the algorithm could choose to conditionalize in deriving the
probabilistic model. For our initial experiments, we utilized a set of
easily computable, but fairly crude, characteristics.\footnote{One
could imagine a variety of more detailed and informative
characteristics of context than those used here.  However, in
performing these experiments, we are interested in how far we can get
with a fairly simple strategy that will port relatively easily to new
domains, rather than relying heavily on information that is 
specific to our current domain.  A fairly coarse-grained set of
characteristics also allows us to restrict ourselves to a relatively
small set of training data; likewise we will not want to encode a
large set of data for each new domain.}  These characteristics fall
into three categories.  In what follows, we take S and T to be
arbitrary templates where the natural language expression from which T
was created appears later in the text than the expression from which S
was created.

The first category relates to the contents of the templates
themselves.  We model the relationship between S and T as one of the
following: S and T have {\it identical slot values}, S is {\it
properly subsumed by} T, S {\it properly subsumes} T, or S and T are
{\it otherwise consistent}.  For instance, in our example in
Section~\ref{problem-section}, template A 
is properly subsumed by template B, and A, B, and C are all properly
subsumed by D, since in each case the latter template is more general
than the former.  We also have a binary characteristic for S and T
having at least two (non-nil) slot values in common.  Finally, we have
a characteristic for modeling when the values of the NAME slot of a
template are both multi-worded and identical; this is a crude
heuristic for identifying matching unique identifiers.

The second category of characteristics relates to the form of
reference used in the expression from which T was created,
specifically whether it was described with an {\it indefinite} phrase,
a {\it definite} phrase (including pronouns), or {\it neither} of
these (e.g., a bare, non-pronominal noun phrase).  In the case of
definite expressions, we also consider the recommendations of a
distinct coreference module within FASTUS.  We have a characteristic 
representing whether the potential antecedent is the {\it preferred}
antecedent,\footnote{Preferred reference is a transitive relation,
that is, template S is treated as a preferred referent of template T
if there is a chain of preferred referents linking them, e.g., if
there is a template R that is the preferred referent of T and template
S is the preferred referent of R.}  a {\it non-preferred, but
possible} antecedent, or not on the list of possible
antecedents.\footnote{Although we do not model information about the
surface positions of the expressions from which S and T were created
within their respective sentences, the coreference module does take
such information into account in determining likely antecedents of
definite expressions.}

The final category of characteristics relates to the distance in the
text between the expressions from which S and T were created, which
we categorize as being in one of five equivalence
classes: {\it very close}, {\it close}, {\it mid-distance}, {\it far
away}, and {\it very far away}. These distances are measured 
crudely (i.e., by character length) so as not to be dependent on the
accuracy of methods for identifying more complex boundaries (e.g.,
clause, sentence, and discourse segment boundaries).

The results of training the maximum entropy models are discussed in
Section~5.  To illustrate the approaches described in the next
section, we will use the probabilities for the templates from the
example passage in Section~\ref{problem-section}, shown in Table~1,
which were produced from the parameters induced from one of the
training sets.

\begin{table}
\begin{center}
\begin{tabular}{|c|c|c|} \hline 
Template S & Template T & Probability \\ \hline \hline
A & B & 0.671  \\ \hline
A & D & 0.505 \\ \hline
B & D & 0.752 \\ \hline
C & D & 0.504 \\ \hline
\end{tabular}
\end{center}
\caption{Pairwise Probabilities for Example Coreference Set}
\label{pairwise-table}
\end{table}

\section{Inferring a Model for Coreference Sets}
\label{coreference-set-section}

We now have a method for obtaining a model that assigns probabilities
to the pairs of templates (henceforth, ``pairwise probabilities'') in
a coreference set that can possibly corefer.  If there are only two
templates in the coreference set, then we have the distribution we
seek.  However, if there are more than two templates, we must utilize
the pairwise probabilities to derive a distribution over the members
of the set of coreference configurations.  In the following sections,
we describe two approaches to recovering such a distribution, followed
by a description of two baseline metrics.  An evaluation of these
approaches is then given in Section~5.

\subsection{An Evidential Reasoning Approach}

\label{dempster-section}

The first approach we describe uses the pairwise probabilities as
sources of evidence that inform the choice of model for the
coreference sets.  The list of coreference configurations for our
example passage are repeated below; we will refer
to these configurations by their corresponding numbers.

\begin{center}
\begin{tabular}{l l} 
1. (A B D) (C)  & 5. (B D) (A) (C) \\
2. (A B) (C D)  &  6. (C D) (A) (B) \\
3. (A B) (C) (D) & 7. (A) (B) (C) (D) \\
4. (A D) (B) (C) & \\
\end{tabular}
\end{center}

We recast a probability that two templates S and T corefer as a
{\it mass distribution} over two members of the power set
of coreference configurations, namely the set containing exactly
those configurations in which S and T occupy the same cell, and the
set containing those in which they do not.  For instance, the
probability that A and B corefer was determined to be 0.671; mapping
this to corresponding sets of coreference configurations results in
the mass distribution $m_{AB}$ in which

\[ m_{AB}(\{\mbox{Configs 1, 2, 3}\}) = 0.671 \]

\noindent and

\[ m_{AB}(\{\mbox{Configs 4, 5, 6, 7}\}) = 0.329 \]
This mass distribution can be seen as representing the beliefs of an
observer who only has access to templates A and B, and who is
therefore ignorant about their relationship to C and D.  We can view
the other pairwise probabilities for the coreference set in the same
manner.

In the best of all worlds, we might identify a model that is
consistent with the mass distributions provided by all the pairwise
probabilities.  However, such a model may not, and often will not,
exist.  This is the case for the pairwise probabilities in our
example, which can be seen most easily by considering only templates
A, C, and D.  The probability of A and D coreferring is 0.505 and of C
and D coreferring is 0.504.  Because we know that A and C cannot
corefer, the coreference configurations in which A and D corefer and
the configurations in which C and D corefer are mutually exclusive.
Therefore, there would have to be a distribution that assigns 0.505 of
probability mass to a set of configurations that is mutually exclusive
from a set that is assigned 0.504 of probability mass.  Obviously,
this cannot be done with a set of probabilities that add up to 1.

This inconsistency arises from the manner in which the pairwise
probabilities are estimated.  The probability of coreference between
templates situated similarly to A and D may be 0.505 with respect to
all contexts in the training data, however it is almost certainly not
this high with respect to the subset of cases in which a template
similar to C is similarly situated.  The same reasoning applies to the
probability of C and D coreferring in light of the existence of A.
Unfortunately, the existence of templates other than the pair being
modeled is the type of conditional information for which we have
little hope of accounting in a general and statistically significant
manner.

Therefore, we may be left with a series of mass distributions defined
over sets of coreference configurations that are in inherent conflict.
Instead of viewing these distributions as constraints on the
underlying probabilistic model, we view them as sources of evidence.
The question is then how to take these sources into account, given
that they may be partially contradictory.  {\it Dempster's Rule of
Combination} \cite{Dempster:68} provides a mechanism for doing this.
Dempster's rule combines two mass distributions $m^1$ and $m^2$ to
form a third distribution $m^3$ that represents the consensus of the
original two distributions; the new mass distribution in effect leans
toward the areas of agreement between the original distributions and
away from points of conflict.  Dempster's rule is defined as follows:

\[ m^{3}(A_{k}) = \frac{1}{1 - \kappa} \sum_{A_{i} \cap A_{j} = A_{k}}
m^{1}(A_{i}) m^{2}(A_{j}) \]
in which 

\[ \kappa = \sum_{A_{i} \cap A_{j} = \emptyset} m^{1}(A_{i})
m^{2}(A_{j}) \] 
The $A_l$ in our case are members of the power set
of possible coreference configurations.  In our
example above, $m_{AB}$ assigns probability mass to two such $A_m$,
the set containing configurations 1, 2, and 3, and the set containing
configurations 4, 5, 6, and~7.

The value $\kappa$ is called the {\it conflict} between the mass
distributions being combined; it provides a measure of the degree of
disagreement between them.  When $\kappa=0$, the original
distributions are compatible; when $\kappa=1$, they are in complete
conflict and the result is undefined.  When $0<\kappa<1$, some
conflict between the distributions exists; Dempster's rule has the
effect of focusing on the agreement between the distributions by
eliminating the conflicting portions and normalizing what remains.

We can therefore use Dempster's Rule to resolve the conflict between
the pairwise probability distributions to generate a distribution over
the coreference configurations.  Because we have pairwise
probabilities for each possibly coreferring pair in the coreference
set, it turns out that the Dempster solution is more easily stated and
computed here than in the general case.  The solution is identical to
the one that results when the probabilities of all the relevant 
pairwise relations (indicating either coreference or not) are
multiplied, normalized by the amount of probability mass
assigned to coreference configurations that are impossible because
coreference is transitive.  For instance, the probability for the
coreference configuration ((A B) (C)) is initially computed to
be\footnote{We use the notation $=_{c}$ to indicate coreference.}

\[p(A =_{c} B)* p(A \neq_{c} C)* p(B \neq_{c} C) \]
However, using this method, impossible combinations (e.g., $A =_{c} B,
B =_{c} C, A \neq_{c} C$) will also receive positive probability mass.
If we normalize the probabilities of possible combinations
by distributing the sum of the probability assigned to all impossible
combinations, the result is the same as that gotten by iteratively
combining the pairwise distributions using Dempster's Rule.

The resulting distribution for our example is: 

\begin{center}
\begin{tabular}{l} 
1. (A B D) (C)  = .383  \\
2. (A B) (C D)  = .184 \\
3. (A B) (C) (D) = .123 \\
4. (A D) (B) (C) = .062 \\
5. (B D) (A) (C)  = .125  \\
6. (C D) (A) (B) = .061 \\
7. (A) (B) (C) (D) = .061 \\
\end{tabular}
\end{center}

In motivating our approach, we noted that we cannot expect to have the
amount of training data necessary to directly estimate distributions
for all the possible scenarios with which we may be confronted.
Limiting ourselves to modeling probabilities between pairs of
templates, however, leads to inconsistencies because of the failure to
take into account the crucial information provided by the existence of
other compatible templates.  Dempster's Rule can be seen as a very
coarse-grained approach to conditioning on context in this regard.
The contributions of the pairwise models are conditioned not on the
existence of other templates in context, but by virtue of the
existence of conflicting models derived from those templates. For
instance, the pairwise probability of coreference between C and D was
originally 0.504, which might be reasonable if those were the only two
templates generated from the text.\footnote{Actually this number is
lower than it would have been, because template B was identified as
the preferred antecedent for template D instead of template C.  If C
and D were the only two templates generated, then C would have been
identified as the preferred antecedent, thus raising the probability.}
However, the probability that C and D corefer in the final
distribution is only 0.245, the sum of the probabilities of the two
partitions in which C and D occupy the same cell.  This adjustment
results from the existence of templates A and B: the fact that
template D has a high probability of coreferring with each, combined
with the fact that template C is incompatible with each, reduces the
likelihood that C and D corefer.  Therefore, the preferences for
particular coreferential dependencies can change when considering the
larger picture of possible coreference sets.

In practice, coreference sets that are significantly larger than the
one we have considered here can lead to an explosive number of
possible coreference configurations.  We have implemented simple
methods for pruning very low probability configurations during
processing and for smoothing the resulting distribution.  The latter
step is accomplished, when necessary, by eliminating certain
low-probability configurations at the end of processing.  The
probability mass from these configurations is distributed uniformly
over all the possible configurations that have been eliminated.  While
this is unlikely to be the best strategy for smoothing from the
standpoint of probabilistic modeling, we are constrained by the number
of alternatives we can report to the downstream system.  Smoothing in
this way allows us to report only the coreference configurations with
non-negligible probability, along with a single probability that is
assigned uniformly to the remainder of the possible configurations.

\subsection{A Model Based on Merging Decisions}

The second approach we consider models the likelihood of correctness
of decisions that a template merger such as the one used in FASTUS
would make in processing a text.  To illustrate, consider the case in
our example in which the probability of the coreference configuration
((A B D) (C)) is determined.  The merger would make the following
decisions in deriving such a configuration, in which the notation
``B\&A'' represents the template that results from templates A and B
having previously been merged.

\begin{center}
\begin{tabular}{l} 
1. B $=_{c}$ A? $\rightarrow$ yes \\
2. C $=_{c}$ B\&A? $\rightarrow$ no \\
3. D $=_{c}$ C? $\rightarrow$ no \\
4. D $=_{c}$ B\&A? $\rightarrow$ yes \\
\end{tabular}
\end{center}

We therefore model the probability of this coreference configuration
as the product of each of the corresponding pairwise probabilities.
Since we cannot model coreference involving objects that have resulted
from  previous (hypothetical) merges -- the appropriate feature values
for distance and form of referring expression would become unclear --
we make the following approximation:

\[ p(X =_{c} Y_1\&...\&Y_n) \approx p(X =_{c} Y_n) \]
in which $Y_n$ is the most recently created template in $Y_1,...,Y_n$.  

Using the probabilities from Table~1,\footnote{We use these
probabilities for ease of comparison.  In reality, the pairwise
probabilities for this model were trained with an adapted set of
training data as explained below, and so these numbers are in
actuality a bit different.} the probability assigned to ((A B D)~(C))
would therefore be

  \[ p(B=_{c}A) * p(C \neq_{c} B) * p(D \neq_{c} C) * p(D
  =_{c} B) = 
  \]

\vspace*{-.25in}

  \[ 0.671 * 1 * (1 - 0.504) * 0.752 = 0.250\] 
Note that unlike the evidential approach, the probability of the pair
D and A coreferring does not come into play, given that coreference
between D and B and between B and A has been factored in.

This approach yields a probabilistic model as given, that is, the
probabilities sum to 1 without normalization.  However, in certain
circumstances the approximation above will generate probability mass
for an impossible case, specifically when it is known a priori that X
is incompatible with one of the templates $Y_1,...,Y_{n-1}$.  For
instance, if templates B and C in our example had been compatible
(with A and C remaining incompatible), then the approximation above
would assign positive probability mass to the coreference
configuration ((A B C)~(D)), because the zero probability of A
coreferring with C would not come into play.  Therefore we modify the
above approximation to apply only if $X$ and each of $Y_1,...,Y_{n-1}$
are compatible; otherwise, the probability mass assigned is used for
normalization.  One can see that this can only improve the pure
form of the model.

Using the pairwise probabilities from Table~1, the results of the
model as applied to the example are:

\begin{center}
\begin{tabular}{l} 
1. (A B D) (C)  = .250  \\
2. (A B) (C D)  = .338 \\
3. (A B) (C) (D) = .083 \\
4. (A D) (B) (C) = .020 \\
5. (B D) (A) (C)  = .123  \\
6. (C D) (A) (B) = .166 \\
7. (A) (B) (C) (D) = .020 \\
\end{tabular}
\end{center}

\subsection{Two Bases of Comparison}

We compared the two learned models with two baseline models.  First,
as an absolute baseline, we compared the model with the uniform
distribution, that is, the distribution that assigns equal probability
to each alternative.  We then sought a more challenging, yet
straightforward baseline.  We defined a simple, ``greedy'' approach to
merging similar to the one used in FASTUS, in which merging of
newly-created templates is attempted iteratively through the prior
discourse, starting with the most recently produced object.  Any
unifications that succeed are performed.  For instance, in the above
example, the greedy method produces the configuration ((A
B) (C D)), because A is compatible with B, C is not compatible with
either, and D is compatible with C (with which merging would be
attempted before the earlier-evoked templates B and A).
Alternatively, in cases in which all of the templates in a coreference
set are pairwise compatible, the greedy method will produce the
configuration in which they are all coreferential.

We then calculated how often this approach yielded the correct results
in each training set.  We distinguished between three values: the
percentage of correctness for coreference sets of cardinality 2 (call
this $p_2$), the percentage for coreference sets of cardinality 3
(call this $p_3$), and the percentage for coreference sets of
cardinality 4 or more (call this $p_{>3}$).  The greedy model was 
defined such that the result of the greedy merging strategy is
assigned the appropriate probability $p_k$, with the remainder of the
probability mass $1-p_k$ distributed uniformly among the remaining
possible alternatives.  (No alternatives were included that were {\it
a priori} known to be impossible due to incompatibilities.)

For instance, in the first training set we describe below, $p_2$=.571,
$p_3$=.652, and $p_{>3}$=.344 (the percentage for the whole training
corpus was $p$=.555). If there are 4 templates, and 10 coreference
configurations are possible, then the answer derived by the greedy
strategy would receive probability .344, and the remaining 9
alternatives would receive probability $\frac{1-.344}{9}=.0729$.  In
the second training set we describe below, $p_2$=.646, $p_3$=.600, and
$p_{>3}$=.345 (the percentage for the whole training corpus was
$p$=.549), and in the third training set, $p_2$=.628, $p_3$=.600, and
$p_{>3}$=.280 (the percentage for the whole training corpus was
$p$=.523).

\section{Experiments}
\label{evaluation-section}

\subsection{Training the Maximum Entropy Models}
\label{training-section}

For reasons described below, we trained separate pairwise probability
models for each of the two approaches.  We ran FASTUS over our
development corpus, 72 texts of which produced coreference data.  The
texts gave rise to 132 coreference sets, and produced characteristics
of context for 647 potential coreference relationships between pairs
of templates.  We created a key by analyzing the texts and entering
the correct coreference relationships.

We created three splits of training and test data.  In the first
split, the training set contained 60 messages, giving rise to 110
coreference sets, and the test set contained 12 messages, giving rise
to 22 coreference sets.  In the second split, the training set
contained 57 messages, giving rise to 102 coreference sets, and the
test set contained 15 messages, giving rise to 30 coreference sets.
The third test set was created by combining the first and second test
sets.  The training set contained 47 messages, giving rise to 88
coreference sets, and the test set contained 25 messages (the first
two test sets overlapped by two messages), which gave rise to 44
coreference sets.

For training the maximum entropy model, only the sets of
characteristics of context for pairwise coreference are relevant; the
number of such sets differed between the two approaches as discussed
below.  The evaluations were performed on the test sets with respect
to the final distribution generated for the coreference sets, with
the result being measured in terms of the average cross-entropy
between the model and the test data.

\paragraph{Data for the Evidential Model}

The evidential model utilizes the pairwise probabilities between all
pairs of templates in a coreference set.  Therefore, we used all such
pairs in each training set to train the maximum entropy model.  In the
first training set, the 110 coreference sets gave rise to
characteristics of context for 578 pairs of templates; in the second,
the 102 coreference sets gave rise to characteristics for 581 pairs of
templates.  In the third training set, the 88 coreference sets gave
rise to characteristics for 525 pairs of templates.

The maximum entropy algorithm selected similar sets of features to
model in each case.\footnote{The following features represent the
referenced characteristic of context paired with the result of
coreference.}  Among the systems of $\lambda_i$ values learned,
negative values were learned for the features in which template S {\it
properly subsumes} template T and in which S and T are {\it otherwise
consistent}.  These two features model the cases in which template T
contains information not contained in template S, reflecting the fact
that expressions referring to the same entity usually do not become
more specific as the discourse proceeds.  A positive value was learned
for the feature modeling cases in which templates S and T had at least
two identical non-nil slot values, as well as for the feature modeling
an exact match of complex name values. As one might expect, a negative
value was learned for the case in which template T was created from an
indefinite expression.  A positive value was learned for the case in
which template T was created from a definite expression and S was
(perhaps transitively) the preferred referent according to the
coreference module.  Interestingly, no value was learned for template
S being a possible but non-preferred referent, but a small positive
value was learned for it not being on the list at all -- presumably
this covers cases in which the coreference module fails to identify an
existing referent.  All the distance features except for {\it close}
and {\it mid-distance} received negative $\lambda_i$ values,
suggesting that coreference between {\it close} and {\it mid-distance}
templates was more likely than coreference between templates that were
{\it very close}, {\it far away}, and {\it very far away}.

The cross-entropy of the learned model as applied to the training data
in each case was about 0.80.  Given that the cross-entropy of the
uniform distribution and the data is 1 (as there are only two possible
values for the random variable, i.e., S and T are coreferent or not),
this relatively small reduction suggests that the problem
has some amount of difficulty, which is consistent with the notable lack
of clear signals of coreference characteristic of the texts in our
domain.

\paragraph{Data for the Merging Decision Model}

Unlike the evidential model, the merging decision model does not
always utilize all of the pairwise probabilities between pairs in a
coreference set.  For instance, in determining the probability of a
coreference configuration ((A B C)), it does not consider the
probability assigned to the pair A and C except to check that they are
compatible.  Therefore, the training set for the maximum entropy
algorithm was pared down to only contain those pairs that the merger
would have considered in deriving the correct coreference
configurations.  The resulting data had the same coreference sets
as the training data for the evidential approach, but consisted of
characteristics of context for 415 template pairs in the first
training set, 405 pairs in the second training set, and 370
pairs in the third training set.  The features selected were
similar to those in the training of the evidential model.

The cross-entropies of the learned maximum entropy models and the
training data were notably better than those for the evidential model,
at about 0.70 in each case.  This improvement is not particularly
surprising.  In the evidential case, the fact that all pairs of
templates are considered results in a certain amount of ``washing
out'' of the data, due to redundancy in coreference relationships.
For instance, coreference between two templates that are far away
might be unlikely if there are no coreferring expressions between
them, but quite likely if there are.  When just considering the
pairwise feature sets, these two cases are not distinguished, so the
resulting probability will be mixed.  However, in the merging decision
case, pairs that are far away will not be in the data set if there are
coreferring expressions between them, and thus the probability for
coreference at long distances will be diminished.  The result is a
``cleaner'' set of data in which clearer distinctions may be found, as
evidenced by the lower cross-entropy achieved.

\subsection{Evaluation Results}

The cross-entropies of the various approaches as applied to the three
sets of test data are shown in Table~2.  The number within parentheses
indicates the number of times that the coreference set with the
highest probability was the correct one.
\begin{table*}
\begin{center}
\begin{tabular}{|c||c|c|c|} \hline 
 & Test Set 1 & Test Set 2 & Test Sets 1 and 2 \\ \hline \hline
Uniform & 2.12 (---) & 1.76 (---) & 2.01 (---) \\ \hline
Greedy & 1.50 (15) & 1.30 (20) & 1.41 (30) \\ \hline
Merging Decision & 1.32 (15) & 1.13 (20) & 1.27 (27) \\ \hline 
Evidential & 1.10 (17) & 0.89 (21) & 1.00 (35) \\ \hline
\end{tabular}
\end{center}
\label{results-table} 
\caption{Initial Evaluation Cross-Entropies}
\end{table*}
As hoped, both the evidential and merging decision approaches
outperformed the uniform and greedy approaches with respect to
cross-entropy.\footnote{The merging decision approach did not do any
better than the greedy approach in terms of raw accuracy, and in fact
did somewhat worse in the third test.  Again, however, the reduction
in cross-entropy is important, as the statistics produced by the
system will be integrated with other probabilistic factors in the
downstream system.}

Interestingly, and perhaps surprisingly, the evidential approach
outperformed the merging decision model, even though in many respects
the latter is more natural and elegant.  While considering feature
sets for all pairs may wash out the training data for the pairwise
probability model somewhat, the evidence provided by all pairs appears
to more than make up for the difference.  Given that a goal of these
experiments is to see how well the strategies would perform with a
fairly crude, easily computable, and portable set of characteristics
of context, we are encouraged by the results of these experiments,
especially considering the limited amount of training data that was
available.

Nonetheless, additional data is necessary to confirm the results of
these initial evaluations.  Although the consistency of the results
between the first two training/test divisions may suggest that the
amount of training data is sufficient for the rather coarsely grained
feature set used, the size of the test sets are potentially of
concern, which motivated our inclusion of the third training/test
division.  Despite the reduction in training data and corresponding
increase in test data, the results of this experiment appear to
consistent with the first two.

There are a variety of characteristics of context that one might
add to improve the models.  For instance, one could
add a characteristic indicating when a template is created from a
phrase in a subject line or table, as many cases of coreference with
subsequent indefinite phrases occur in this circumstance.  Other types
of information about text type, text structure, and more
finely grained distinctions with respect to referential types (e.g.,
modeling pronouns differently than other definite NPs) would all
likely further improve the model, although for some of these
additional training data would be required and more domain and genre
dependence may result.

While this work was motivated by a need to pass probabilistic output
to a downstream data fusion system, these methods can be applied
system internally also, to supplant existing algorithms for merging in
IE settings that do not allow for probabilistic output.  In this
scenario, the system simply performs the template merging dictated by
the most probable coreference configuration for a given coreference
set.  However, as noted earlier, the texts in our application are
relatively short, and therefore the coreference sets are usually of
manageable size.  Significantly larger coreference sets can lead to an
enormous number of possible coreference configurations.  Therefore, to
address this task in applications with much longer texts, mechanisms
beyond those that were necessary here will be required for
intelligently pruning the search space and subsequently smoothing the
distributions.

\section{Conclusions}

Certain applications require that the output of an information
extraction system be probabilistic, so that a downstream system can
reliably {\it fuse} the output with possibly contradictory information
from other sources.  In this paper we considered the problem of
assigning a probability distribution to alternative sets of
coreference relationships among entity descriptions.  We presented the
encouraging results of initial experiments with several approaches to
estimating such distributions in an application using SRI's FASTUS
information extraction system.  We would expect further gains from
encoding additional training data and modeling more informative
characteristics of context.

\section*{Acknowledgments}

The author thanks John Bear, Joshua Goodman, and two anonymous
reviewers for helpful comments and criticisms, and the SRI Message
Handler project team for their contributions to the system in which
this work is embedded.  This work was supported by the Defense
Advanced Research Projects Agency under contract number 4099SCL001
(E-Systems Inc., prime contractor).

\end{document}